\documentclass[11pt,a4paper]{article}
\usepackage[hyperref]{acl2021}
\usepackage{times}
\usepackage{latexsym}
\usepackage{algorithmic}
\usepackage{algorithm}
\usepackage{graphicx}
\graphicspath{ {./images/} }

\usepackage{microtype}

\aclfinalcopy 

\title{Enhancing Transformers with Gradient Boosted Decision Trees for NLI Fine-Tuning}

\author{Benjamin Minixhofer{\normalfont \textsuperscript{1}}\thanks{\,\,\,Work conducted as Research Intern at Huawei Noah's Ark Lab.}  {\normalfont ,} Milan Gritta{\normalfont \textsuperscript{2} and }Ignacio Iacobacci{\normalfont \textsuperscript{2}}\\
  \textsuperscript{1}Johannes Kepler University, Linz, Austria \\
  \textsuperscript{2}Huawei Noah’s Ark Lab, London, UK\\
  \texttt{bminixhofer@gmail.com}\\
  \texttt{\{milan.gritta, ignacio.iacobacci\}@huawei.com} \\}

\date{}

\begin{document}
\maketitle
\begin{abstract}
Transfer learning has become the dominant paradigm for many natural language processing tasks. In addition to models being pretrained on large datasets, they can be further trained on intermediate (supervised) tasks that are similar to the target task. For small Natural Language Inference (NLI) datasets, language modelling is typically followed by pretraining on a large (labelled) NLI dataset before fine-tuning with each NLI subtask.
In this work, we explore Gradient Boosted Decision Trees (GBDTs) as an alternative to the commonly used Multi-Layer Perceptron (MLP) classification head. GBDTs have desirable properties such as good performance on dense, numerical features and are effective where the ratio of the number of samples w.r.t the number of features is low.
We then introduce FreeGBDT, a method of fitting a GBDT head on the features computed during fine-tuning to increase performance without additional computation by the neural network. We demonstrate the effectiveness of our method on several NLI datasets using a strong baseline model (RoBERTa-large with MNLI pretraining). The FreeGBDT shows a consistent improvement over the MLP classification head.
\end{abstract}

\section{Introduction}
\label{introduction}

Recent breakthroughs in transfer learning ranging from semi-supervised sequence learning \cite{dai2015semisupervised} to ULMFiT \cite{howard2018universal}, ELMo \cite{peters2018deep} and BERT \cite{devlin2018bert} have brought significant improvements to many natural language processing (NLP) tasks. Transfer learning involves pretraining neural networks, often based on the Transformer \cite{vaswani2017attention}, on large amounts of text in a self-supervised manner in order to learn transferable language features useful for many NLP tasks. Pretraining is followed by fine-tuning the model on the target task. Pretrained models can also be further trained on intermediate labelled datasets which are similar to the target task before the final fine-tuning stage \cite{pruksachatkun-etal-2020-intermediate}. We refer to this as \textit{intermediate supervised pretraining}. 
In this manner, the network learns more meaningful internal representations of the input text that are better aligned with the target task. In order to fine-tune the pretrained network, some latent representation of the text (e.g. the hidden state corresponding to the \verb|[CLS]| token of BERT-like models) is used as the input to a \textit{classification head}, usually a randomly initialised Multi-Layer Perceptron (MLP) \cite{wolf2019huggingfaces}. The output of the classification head can then be interpreted as probability distribution over classes. The input to the classification head is referred to as \textit{features} throughout the paper. It is a high-dimensional vector that serves as a rich, distributed representation of the input text.\\


We investigate whether replacing the commonly used MLP classification head with a GBDT \cite{friedman2001greedy} can provide a consistent improvement, using NLI tasks as our use case. GBDTs are known for strong performance on dense, numerical features \cite{ke2019deepgbm}, which includes the hidden states in a neural network. The number of input features $p$, i.e. the dimension of the hidden state corresponding to the \verb|[CLS]| token is not necessarily much larger than the number of samples $n$ (it may even be smaller). GBDTs have proven effective for tasks where $n < p$ \cite{kong2018deep} and can be more effective compared to logistic regression if $n \not\gg p$ \cite{couronne2018random}. Therefore, for a language model that was trained on an intermediate supervised task before fine-tuning, we hypothesise that a GBDT may be able to outperform an MLP classification head as the hidden states already encode information relevant to the target task at the start of fine-tuning. The head must learn to exploit this information exclusively during the fine-tuning stage in which the training data may consist of only a few samples. Our contributions are as follows:

\begin{itemize}
  \item We integrate the GBDT into a near state-of-the-art (SOTA) language model as an alternative to an MLP classification head and train on the features extracted from the model \textit{after} fine-tuning. We refer to it as standard GBDT.
  \item We introduce a method to train a GBDT on the features computed \textit{during} fine-tuning, at no extra computational cost by the neural network, showing a consistent improvement over the baseline. We refer to it as \textbf{FreeGBDT}.
\end{itemize}

In the following, we recap different approaches to integrating tree-based methods with neural networks (Section \ref{related_work}). We introduce our FreeGBDT method in Section \ref{methodology}. We present our experimental setup in Section \ref{experimental_setup} and results on standard NLI benchmarks in Section \ref{results_analysis}. To conclude, we discuss improvements and limitations of our method in Section \ref{discussion}.

We release our code\footnote{Code will be available at \url{https://github.com/huawei-noah/free-gbdt}.}, implemented with LightGBM \cite{ke2017lightgbm} and Huggingface's Transformers \cite{wolf2019huggingfaces}, to the NLP community.

\section{Related Work}
\label{related_work}

Recent work on transfer learning in NLP has often been based on pretrained transformers, e.g. BERT \cite{devlin2018bert}, XLNet \cite{yang2019xlnet}, T5 \cite{raffel2019exploring} and RoBERTa \cite{liu2019roberta}. These models are pretrained on large datasets using self-supervised learning, typically a variation of language modelling such as Masked Language Modelling (MLM). MLM consists of masking some tokens as in the Cloze task \cite{taylor1953cloze}. The objective of the model is to predict the masked tokens. Recently, approaches using alternatives to MLM such as Electra \cite{clark2020electra} and Marge \cite{lewis2020pre} have also been proposed. Pretraining transformers on large datasets aims to acquire the semantic and syntactic properties of language, which can then be used in downstream tasks. The models can additionally be trained in a supervised manner on larger datasets before being fine-tuned on the target task.

\paragraph{Natural Language Inference} is one of the most canonical tasks in Natural Language Understanding (NLU) \cite{nie2019adversarial,bowman2015large}. NLI focuses on measuring \textit{commonsense reasoning} ability \cite{10.1145/2701413} and can be seen as a proxy task that estimates the amount of transferred knowledge from the self-supervised phase of training. The task involves providing a \textit{premise} (also called \textit{context}) and a \textit{hypothesis} that a model has to classify as:

\begin{itemize}
    \item \textit{Entailment}. Given the context, the hypothesis is correct.
    \item \textit{Contradiction}. Given the context, the hypothesis is incorrect.
    \item \textit{Neutral}. The context neither confirms nor disconfirms the hypothesis.
\end{itemize}

The task can also be formulated as binary classification between \textit{entailment} and \textit{not\_entailment} (\textit{contradiction} or \textit{neutral}). We focus on NLI as a challenging and broadly applicable NLP task, with multiple smaller evaluation datasets being available as well as the large Multi-Genre Natural Language Inference corpus \cite[MNLI]{N18-1101}, which is often used for effective intermediate pretraining \cite{liu2019roberta}. As such, it provides a testing ground for the GBDT classification head with intermediate supervised pretraining.

\paragraph{Tree-based methods} Models based on decision trees have a long history of applications to various machine learning problems \cite{breiman1984classification}. Ensembling multiple decision trees via bagging \cite{breiman1996bagging} or boosting \cite{freund1999short} further improves their effectiveness and remains a popular method for modelling dense numerical data \cite{feng2018multi}. Ensemble methods such as Random Forests \cite{breiman2001random} and GBDTs combine predictions from many weak learners, which can result in a more expressive model compared to an MLP. There have been several approaches to combining neural networks with tree-based models, approximately divided into two groups.

\begin{enumerate}
  \item \textbf{Heterogeneous ensembling}: The tree-based model and the neural network are trained independently, then combined via ensembling techniques. \textit{Ensembling} refers to any method to combine the predictions of multiple models such as \textit{stacking} \cite{wolpert1992stacked} or an arithmetic mean of the base models' predictions.
  \item \textbf{Direct integration}: The tree-based model is jointly optimised with the neural network.
\end{enumerate}

Heterogeneous ensembling \cite{li2019combining} has proven effective for many applications such as Online Prediction \cite{ke2019deepgbm}, Learning-to-Rank for Personal Search \cite{lazri2018combination} and Credit Scoring \cite{xia2018novel}. It is also suitable for multimodal inputs, e.g. text, images and/or sparse categorical features as some input types are better exploited by a neural network while others are amenable to tree-based models \cite{ke2019deepgbm}.

\begin{figure}[t!]
\includegraphics[width=7.5cm]{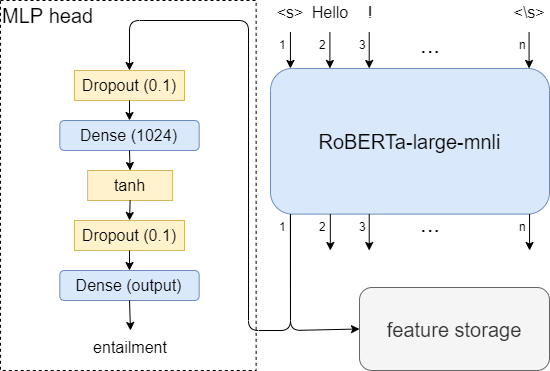}
\caption{The baseline model architecture. Feature storage is populated \textit{during} fine-tuning for the FreeGBDT but \textit{after} fine-tuning for the standard GBDT method.}
\label{fig:model}
\end{figure}

Direct integration makes the tree-based model compatible with back-propagation thus trainable with the neural network in an end-to-end manner. Examples include the Tree Ensemble Layer \cite{hazimeh2020tree}, Deep Neural Decision Forests \cite{kontschieder2015deep} and Deep Neural Decision Trees \cite{yang2018deep}. Deep Forests \cite{zhou2017deep} are also related although they aim to create deep non-differentiable models instead. Other examples include driving neural network fine-tuning through input perturbation \cite{bruch2020learning}, which focuses specifically on using a tree ensemble to fine-tune the neural network representations.

As pretrained transformer-based models have recently achieved strong performance on various NLP tasks \cite{devlin2018bert}, we see an opportunity to take advantage of their distributed representations by the means of using a tree-based model as the classification head. Our methods differ from direct integration in that they are not end-to-end differentiable. The training procedure is a sequence, i.e. the transformer-based model is fine-tuned first, then a GBDT is trained with features extracted from the model. We do not interfere with the model updates during training. Finally, the GDBT replaces the MLP classification head. Our approach is invariant to the method with which the neural network is fine-tuned as long as there exists a forward pass in which the features are computed. Recent methods for neural network fine-tuning include FreeLB \cite{zhu2019freelb} and SMART \cite{jiang2019smart}. FreeLB is an adversarial method, which perturbs the input during training via gradient ascent steps to improve robustness. SMART constrains the model updates during fine-tuning with smoothness-inducing regularisation in order to reduce overfitting. These approaches could theoretically be combined with both the standard GBDT and the FreeGBDT.

\section{Methodology}
\label{methodology}

We introduce the standard GBDT (Algorithm \ref{alg:standard}) and the FreeGBDT (Algorithm \ref{alg:free}), our new method of using features generated during fine-tuning. \textit{Features} refers to the hidden state corresponding to the \verb|[CLS]| token of BERT-like pretrained models. We use these features as training data for the GBDT and FreeGBDT.

\subsection{The standard GBDT classification head}
\begin{figure}[t!]
\includegraphics[width=7.5cm]{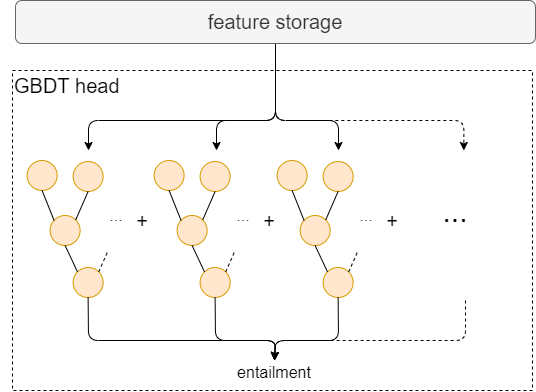}
\centering
\caption{The GBDT classification head.}
\label{fig:gbdt}
\end{figure}

In order to train the standard GBDT, we apply the feature extraction procedure shown in Algorithm \ref{alg:standard}. Using the fine-tuned neural network, we perform one additional forward pass over each sample in the training data. We store the features as training data for the GBDT, denoted 'feature storage' in Figures \ref{fig:model} and \ref{fig:gbdt}. The GBDT can be then used as a substitute for the MLP classification head.

\begin{algorithm}[t!]
\caption{Standard GBDT training procedure. Features are extracted after fine-tuning.}
\label{alg:standard}
\begin{algorithmic}
\REQUIRE{training data $X$, pretrained network $f_p$ parametrised by $\theta_p$, classification head $f_h$ parametrised by $\theta_h$.}
\FOR{$epoch=1..N_{epochs}$}
    \FOR{minibatch $(X_b, y_b) \subset X$}
        \STATE{$cls\gets f_p(X_b, \theta_p)$}
        \STATE{$y_{pred}\gets f_h(cls, \theta_h)$}
        \STATE{$loss\gets LossFn(y_{pred}, y_b)$}
        \STATE{update $\theta_h$ and $\theta_p$ via backpropagation of $loss$}
    \ENDFOR
\ENDFOR
\STATE
\STATE{$features\gets \textrm{empty list}$}
\STATE{$labels\gets \textrm{empty list}$}
\STATE
\FOR{minibatch $(X_b, y_b) \subset X$}
    \STATE{$cls\gets f_p(X_b, \theta_p)$}
    \STATE{extend $features$ with $cls$}
    \STATE{extend $labels$ with $y_b$}
\ENDFOR

\STATE
\STATE{$gbdt\gets train_{gbdt}(features, labels)$}

\end{algorithmic}
\end{algorithm}

\subsection{The FreeGBDT classification head}

Instead of extracting features once after fine-tuning, the training data for the proposed FreeGBDT is obtained \textit{during} fine-tuning. The features computed in every forward pass of the neural network are stored as training data, shown in Algorithm \ref{alg:free}. As no additional computation by the neural network is required, this new classification head is called \textit{FreeGBDT}. Accumulating features in this manner allows the FreeGBDT to be trained on $N \times E$ samples while the standard GBDT is trained on $N$ samples where $N$ is the size of the dataset and $E$ is the number of fine-tuning epochs.

\begin{table}[h]
\centering
\begin{tabular}{lrrrc}
\hline
\textbf{Corpus} & \textbf{Train} & \textbf{Dev} & \textbf{Test} & \textbf{Classes} \\
\hline
ANLI & 162k & 3.2k & 3.2k & 3 \\
CNLI & 6.6k & 800 & 1.6k & 3 \\
RTE & 2.5k & 278 & 3k & 2 \\
CB & 250 & 57 & 250 & 3 \\
QNLI & 104k & 5.4k & 5.4k & 2 \\
\hline
\end{tabular}
\caption{NLI evaluation datasets. The tasks with 3 classes contain labels: \textit{entailment}, \textit{neutral} and \textit{contradiction}, tasks with 2: \textit{entailment} and \textit{not\_entailment}.}
\label{tab:tasks}
\end{table}

\section{Experimental Setup}
\label{experimental_setup}

We now describe the featured models, details of training procedures and evaluation datasets.

\subsection{Datasets}

\begin{algorithm}[t!]
\caption{FreeGBDT training procedure. Features are accumulated throughout fine-tuning.}
\label{alg:free}
\begin{algorithmic}
\REQUIRE{training data $X$, pretrained network $f_p$ parametrised by $\theta_p$, classification head $f_h$ parametrised by $\theta_h$.}
\STATE{$features\gets \textrm{empty list}$}
\STATE{$labels\gets \textrm{empty list}$}
\STATE
\FOR{$epoch=1..N_{epochs}$}
    \FOR{minibatch $(X_b, y_b) \subset X$}
        \STATE{$cls\gets f_p(X_b, \theta_p)$}
        \STATE{extend $features$ with $cls$}
        \STATE{extend $labels$ with $y_b$}
        \STATE{$y_{pred}\gets f_h(cls, \theta_h)$}
        \STATE{$loss\gets LossFn(y_{pred}, y_b)$}
        \STATE{update $\theta_h$ and $\theta_p$ via backpropagation of $loss$}
    \ENDFOR
\ENDFOR
\STATE
\STATE{$gbdt\gets train_{gbdt}(features, labels)$}
\end{algorithmic}
\end{algorithm}

We evaluate our methods on the following NLI datasets, summarised in Table \ref{tab:tasks}.

\begin{table*}[h]
\setlength\tabcolsep{7.7pt}
\centering
\begin{tabular}{lccccc}
\hline
\textbf{Method} & \textbf{CB} & \textbf{RTE} & \textbf{CNLI} & \textbf{ANLI} & \textbf{QNLI} \\
\hline
MLP head &   93.57 (2.2) &  89.51 (0.8) & 82.49 (0.8) & 57.56 (0.5) &  \textbf{94.32 (0.1)} \\
standard GBDT &  94.11 (1.7) &  89.33 (0.8) & 80.84 (0.9) & 57.22 (0.5) &  94.29 (0.1) \\
\hline
FreeGBDT &   \textbf{94.20 (1.7)} &  \textbf{89.69 (0.8)} & \textbf{82.53 (0.7)} &  \textbf{57.63 (0.5)} &  94.30 (0.1) \\
\hline
\end{tabular}
\caption{Mean Accuracy (Standard Deviation) on the development sets from 20 runs with different random seeds. A Wilcoxon signed-rank test conducted across all five datasets confirms significance with $p \approx 0.01$ c.f. Section \ref{results_analysis}.}
\label{tab:results}
\end{table*}

\begin{itemize}
    \item Adversarial NLI (ANLI) \cite{nie2019adversarial}. This corpus consists of three rounds of data collection. In each round, annotators try to break a model trained on data from previous rounds. We use the concatenation of R1, R2 and R3.

    \item Counterfactual NLI (CNLI) \cite{kaushik2019learning}. The CNLI corpus consists of counterfactually-revised samples of SNLI \cite{bowman2015large}. We use the full dataset i.e. samples with the revised premise and with the revised hypothesis.
    
    \item Recognising Textual Entailment (RTE) \cite{wang2018glue}. We use the data and format as used in the GLUE benchmark: a concatenation of RTE1 \cite{dagan2006pascal}, RTE2 \cite{bar2006second}, RTE3 \cite{giampiccolo2007third} and RTE5 \cite{bentivogli2009fifth}, recast as a binary classification task between \textit{entailment} and \textit{not\_entailment}.
   
    \item CommitmentBank (CB) \cite{de_Marneffe_Simons_Tonhauser_2019}. We use the subset of the data as used in SuperGLUE \cite{wang2019superglue}.
   
    \item Question-answering NLI (QNLI) \cite{demszky2018transforming}. This is a converted version of the Stanford Q\&A Dataset \cite{rajpurkar2016squad}, aiming to determine whether a given context contains the answer to a question.
\end{itemize}

\subsection{Model and Training}

We start all experiments from the RoBERTa-large model \cite{liu2019roberta} with intermediate pretraining on the Multi-Genre Natural Language Inference (MNLI) corpus \cite{N18-1101}. The MNLI checkpoint is provided by the fairseq\footnote{\url{https://github.com/pytorch/fairseq}} library \cite{ott2019fairseq}. 
Note that no task-specific tuning of hyperparameters was performed. Instead, we use one learning rate cycle \cite{smith2017cyclical} with a maximum learning rate of $1 \times 10^{-5}$ for each task to fine-tune RoBERTa for 10 epochs with a batch size of 32. We use the Adam optimiser \cite{kingma2014adam} to optimise the network. In order to compare the FreeGBDTs with standard GBDTs, we apply Algorithm \ref{alg:standard} and Algorithm \ref{alg:free} during the same fine-tuning session to eliminate randomness from different model initialisations.\\

We use LightGBM\footnote{\url{https://lightgbm.readthedocs.io}} to train the GBDT. We do not manually shuffle the data before training. The individual trees of a GBDT are learned in a sequence where each tree is fit on the residuals of the previous trees. One important parameter of the GBDT is thus the number of trees to fit. This is commonly referred to as \textit{boosting rounds}.

\begin{table}[h]
\centering
\setlength\tabcolsep{2.2pt}
\begin{tabular}{lllllll}
\hline
\textbf{Parameter} & \textbf{CB} & \textbf{RTE} & \textbf{CNLI} &  \textbf{ANLI} & \textbf{QNLI} \\
\hline
learning rate & 0.1 & 0.1 & 0.1 & 0.1 & 0.1 \\
max. leaves &  256 & 256 & 256 & 256 & 256 \\
boosting rounds & 10 & 10 & 10 & 40 & 30 \\
\hline
\end{tabular}
\caption{Hyperparameters of the standard GBDT and the FreeGBDT. All other hyperparameters are set to the default value as per LightGBM version 2.3.1.}
\label{tab:tree_params}
\end{table}

We observe that the optimal number of boosting rounds varies significantly across tasks, with a tendency towards more boosting rounds for larger datasets. Thus, we select the number of boosting rounds from the set $\{ 1, 10, 20, 30, 40 \}$ for each task. This is the only task-specific hyperparameter in our experiments. The hyperparameters are identical for the standard GBDT and the FreeGBDT, shown in Table \ref{tab:tree_params}, the model shown in Figure \ref{fig:gbdt}.\\
The time it takes to train the standard GBDT and the FreeGBDT is negligible compared to the time it takes to fine-tune the RoBERTa model.

\begin{figure*}[t]
\includegraphics[width=16cm]{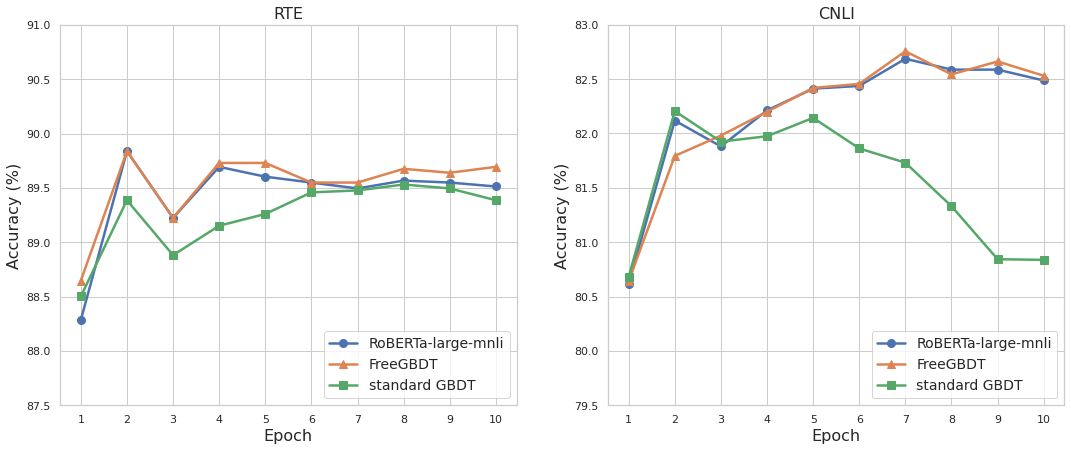}
\caption{Accuracy on RTE and CNLI development sets. Training is paused after each epoch to compare the GBDT, FreeGBDT and MLP heads. We plot the mean from 20 runs (same hyperparameters but different seeds).}
\centering
\label{fig:rte_training}
\end{figure*}

\subsection{Evaluation}

\paragraph{Development set} We evaluate our methods using accuracy. Each experiment is repeated 20 times with different random seeds. We report the mean and standard deviation.

\paragraph{Test set} For each task, we train the GBDT with the following boosting rounds: $\{ 1, 10, 20, 30, 40 \}$. We select the GBDT with the best score on the development set. Test scores are obtained with a submission to the SuperGLUE benchmark\footnote{\url{https://super.gluebenchmark.com/}} for CB and the GLUE benchmark\footnote{\url{https://gluebenchmark.com/}} for RTE and QNLI. We calculate the test scores on ANLI and CNLI ourselves as the test labels are publicly available. We report accuracy on the test set for each task except for CB, where we report the mean of F1 Score and Accuracy, same as the SuperGLUE leaderboard.

\section{Results and Analysis}
\label{results_analysis}

We summarise the results on the development sets in Table \ref{tab:results}. The FreeGBDT is compared with a standard GBDT and the MLP classification head. The standard GBDT achieves a higher score than the MLP head on 1 out of 5 tasks. The FreeGBDT outperforms the standard GBDT on 5 out of 5 tasks and the MLP head on 4 out of 5 tasks. As recommended for statistical comparison of classifiers across multiple datasets \cite{JMLR:v7:demsar06a} we conduct a Wilcoxon signed-rank test \cite{wilcoxon1992individual} with the accuracy differences between the FreeGBDT and the MLP across the 20 seeds and 5 datasets. The test confirms that the improvement from our FreeGBDT method is significant with $p \approx 0.01$.\\

\begin{table}[h]
\setlength\tabcolsep{3pt}
\centering
\begin{tabular}{lccccc}
\hline
\textbf{Method} & \textbf{CB} & \textbf{RTE} & \textbf{CNLI} & \textbf{ANLI} & \textbf{QNLI} \\
\hline
MLP head & 92.9 & 87.5 & 83.6 & 57.4 &  \textbf{94.3} \\
GBDT &  91.3 &  87.5 &  82.1 &  57.2 &  \textbf{94.3} \\
\hline
FreeGBDT & \textbf{93.3} & \textbf{87.8} & \textbf{83.7} & \textbf{57.6} & \textbf{94.3} \\
\hline
\end{tabular}
\caption{Results on the test sets.}
\label{tab:test_results}
\end{table}

Results on the test sets are shown in Table \ref{tab:test_results}. The FreeGBDT achieves a small but consistent improvement over the MLP head on each task except QNLI. This task is not a conventional NLI task but a question-answering task converted to an NLI format \cite{demszky2018transforming}. It has been shown that QNLI does not benefit from MNLI pretraining \cite{liu2019roberta} hence this result is not unexpected. Out of the four datasets which do benefit from MNLI pretraining, the FreeGBDT improves over the MLP head on each one with an average score difference of +$0.23\%$. As our experiments start from a competitive baseline, RoBERTa-large with MNLI pretraining, we consider the results important because (a) to the best of our knowledge, this is the first tree-based method that achieves near state-of-the-art performance on benchmark NLI tasks and (b) our method is 'free' as it requires no additional computations by the model. We were able to demonstrate that a FreeGBDT head can be successfully integrated with modern transformers and is a good alternative to the commonly used MLP classification head.\\  

\begin{table*}[h]
\setlength\tabcolsep{7pt}
\centering
\begin{tabular}{lcccc}
\hline
\textbf{Method} &    \textbf{A1} & \textbf{A2} & \textbf{A3} & \textbf{All} \\
\hline
RoBERTa-large-mnli \textit{(ours)} &  72.0 &  52.5 &  49.4 &  57.4 \\
RoBERTa-large-mnli + GBDT \textit{(ours)}    &  72.1 &  52.2 &  49.0 &  57.2 \\
\hline
SMART \cite{jiang2019smart} & 72.4 & 49.8 & \textbf{50.3} & 57.1 \\
ALUM \cite{liu2020adversarial} & 72.3 & 52.1 & 48.4 & 57.0 \\
InfoBERT \cite{wang2020infobert} & \textbf{75.0} & 50.5 & 49.8 & \textbf{58.3} \\
\hline
RoBERTa-large-mnli + FreeGBDT \textit{(ours)} &  71.9 &  \textbf{52.7} &  49.7 &  57.6 \\
\hline
\end{tabular}
\caption{Accuracy across different rounds of the ANLI test set. \textit{All} denotes a sample-weighted average. Our FreeGBDT achieves SOTA on the A2 subset. \allowbreak InfoBERT \cite{wang2020infobert} is the SOTA on the full test set.}
\label{tab:anli_results}
\end{table*}

For the Adversarial NLI (ANLI) dataset, we report results which are competitive with the state-of-the-art shown in Table \ref{tab:anli_results}, surpassing both SMART \cite{jiang2019smart} and ALUM \cite{liu2020adversarial}. The RoBERTa-large model pretrained with SNLI, MNLI, FEVER \cite{thorne2018fever} and ANLI reported $53.7\%$ accuracy on the ANLI test set \cite{nie2019adversarial}. The state-of-the-art result of $58.3\%$ accuracy on the ANLI dataset was achieved by InfoBERT \cite{wang2020infobert}. The FreeGBDT achieves a new state-of-the-art on the A2 subset of ANLI with $52.7\%$. Interestingly, it does not yield an improvement on the easier A1 subset but compares favourably to other recent approaches on the more difficult A2 and A3 subsets of ANLI.

To better understand how performance of the GBDTs evolves during fine-tuning, we carry out an additional experiment. We pause training after each epoch to extract features and train a standard GBDT. We compare it with the MLP classification head and a FreeGBDT trained on the features accumulated up to the current epoch. The result on the RTE and CNLI datasets is shown in Figure \ref{fig:rte_training}. Notably, the FreeGBDT does not improve on the standard GBDT after the first epoch where the number of instances the GBDT is trained on is equal to the size of the training dataset for both. As the FreeGBDT starts accumulating more training data, however, it consistently outperforms the standard GBDT and eventually, the MLP head.\\ 

The state-of-the-art in NLI provides some context for our method of combining tree-based models with modern neural networks. RoBERTa (large with MNLI pretraining) reports $89.5\%$ accuracy on the development set of RTE \cite{liu2019roberta} and $94.7\%$ accuracy on the development set of QNLI. The same model obtains an F1 Score / Accuracy of $90.5/95.2$ on the CB test set and an accuracy of $88.2\%$ on the RTE test set. Note that ensembles of 5 to 7 models \cite{liu2019roberta} were used while our test figures achieve similar scores of $91.3/95.2$ for CB and $87.8\%$ for RTE with a \textit{single} model. These are not direct comparisons, however, the figures demonstrate that FreeGBDT can operate at SOTA levels while matching and exceeding the 'default' MLP head classifier accuracy. Across all datasets, the FreeGBDT improves by an average of $0.2\%$ and $0.5\%$ over the MLP head and the standard GBDT head, respectively.

\section{Discussion}
\label{discussion}

The FreeGBDT improves over the MLP head on each task where intermediate supervised pretraining on MNLI is effective. The improvement is significant but not large. This is expected since the input features of the classification head are already a highly abstract representation of the input. Thus, there is limited potential for improvement. However, our results show that a tree-based method is a viable alternative to the commonly used MLP head and can improve over a baseline chosen to be as competitive as possible. Notably, the FreeGBDT improves the MLP baseline on the CB dataset by $>0.6\%$ solely by switching to our tree-based classification head.\\

Furthermore, the FreeGBDT outperforms a standard GBDT by a large margin in some cases. For instance, we observe a +$1.5\%$ improvement on the CNLI dataset. Figure \ref{fig:rte_training} shows the gap forming towards the end of training. We think this may be due to overfitting to the training data. Recall that the standard GBDT is trained only on features extracted \textit{after} fine-tuning. At this point, the features may exhibit a higher degree of memorisation of the training data. The FreeGBDT is able to mitigate this problem as it was trained with features collected throughout training. Let $f(x,\theta_t)$ denote a mapping from the input text $x$ to the output space parameterised by $\theta_t$ where $t \in \{0..T\}$ and $T$ is the total amount of steps the model is fine-tuned for. Then, the standard GBDT is trained on features from $f(x,\theta_T)$, while the FreeGBDT is trained on features from every $t$ in $\{0..T\}$. As such, it may help to think of the FreeGBDT as a type of regularisation through data augmentation (from the FreeGBDT's point of view), having trained on several perturbed views of each training instance.\\

\begin{figure*}[t]
\includegraphics[width=16cm]{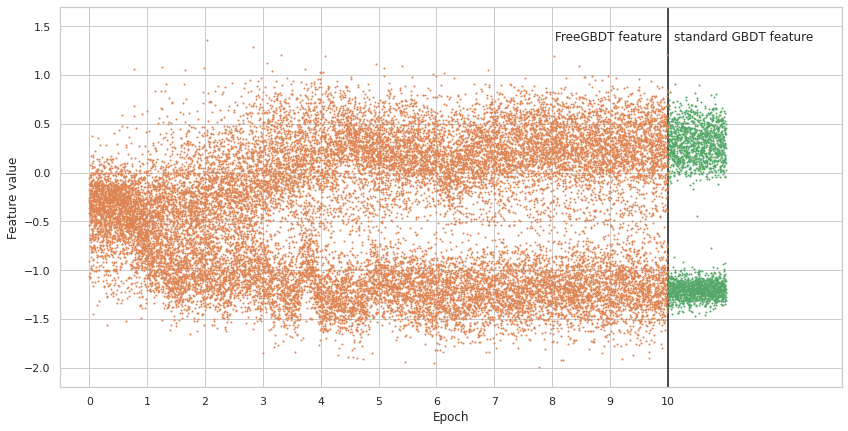}
\caption{Values of a typical dimension from the 1,024 dimensional vector stored \textit{during} fine-tuning to train a FreeGBDT (left). The same dimension extracted \textit{after} fine-tuning to train a standard GBDT (right).}
\label{fig:features_training}
\centering
\end{figure*}

Figure \ref{fig:features_training} helps illustrate the regularisation effect by showing the differences between the FreeGBDT and standard GBDT training data beyond just size. The figure shows the temporal changes in a typical feature collected during fine-tuning and the same feature extracted after fine-tuning. We can see that the distribution gradually changes from earlier epochs but remains similar to the distribution of the feature at the end of fine-tuning. FreeGBDT is able to exploit the information at the start of fine-tuning as the features at $t=0$ already encode information highly relevant to the target task hence all training data is useful. The FreeGBDT head compares favourably to an MLP head, which is randomly initialized at the start of the fine-tuning stage and must thus learn to exploit the latent information from a potentially small amount of training examples. Therefore, we believe intermediate supervised pretraining is essential for the effectiveness of the FreeGBDT, supported by the results from preliminary experiments on the BoolQ dataset \cite{clark2019boolq} and QNLI, which does not benefit from pretraining on MNLI \cite{liu2019roberta} where FreeGBDT matches the accuracy of the MLP head but does not exceed it. Our experiments also suggest that the potential for improvement from FreeGBDT depends on the size of the training dataset. The gap between FreeGBDT and the MLP head in Table \ref{tab:test_results} is larger for the smaller datasets CB and RTE and smaller for the larger datasets (ANLI, CNLI, QNLI). This is consistent with prior work showing that GBDTs are especially effective compared to other methods if $n \not\gg p$ \cite{couronne2018random} and hints that the FreeGBDT method might be especially useful for smaller datasets.

\section{Future Work}
\label{future_work}

One possible avenue for future work is exploring different features to train the GBDT, e.g. the hidden states from different layers of the pretrained model. This includes new combinations of top layer representations of the Transformer to generate richer input features for the classification head. This could lead to potential improvement by leveraging a less abstract representation of the input. Given that our method operates on distributed representations from a pretrained encoder, applications in other domains such as Computer Vision may be possible, e.g. using features extracted from a ResNet \cite{he2016deep} encoder. Furthermore, a GBDT might not be the best choice for each task hence the use of Random Forests \cite{breiman2001random} or Support Vector Machines \cite{cortes1995support} may also be evaluated to investigate the effectiveness of combining Transformer neural networks with traditional supervised learning methods.

\section{Conclusion}
\label{conclusion}

State-of-the-art transfer learning methods in NLP are typically based on pretrained transformers and commonly use an MLP classification head to fine-tune the model on the target task. We have explored GBDTs as an alternative classification head due to their strong performance on dense, numerical data and their effectiveness when the ratio of the number of samples w.r.t the number of features is low. We have shown that tree-based models can be successfully integrated with transformer-based neural networks and that the free training data generated during fine-tuning can be leveraged to improve model performance with our proposed FreeGBDT classification head. Obtaining consistent improvements over the MLP head on several NLI tasks confirms that tree-based learners are relevant to state-of-the-art NLP.

\bibliographystyle{acl_natbib}
\bibliography{acl2021}


\end{document}